\documentclass[10pt,twocolumn,letterpaper]{article}

\usepackage[pagenumbers]{cvpr} 
\usepackage{bm}

\definecolor{cvprblue}{rgb}{0.21,0.49,0.74}
\usepackage[pagebackref,breaklinks,colorlinks,allcolors=cvprblue]{hyperref}
\usepackage{colortbl}
\usepackage{ragged2e}
\usepackage{pifont}
\usepackage[dvipsnames,svgnames,x11names,table]{xcolor}

\definecolor{lightyellow}{named}{LightYellow}
\definecolor{lightred}{RGB}{255,204,203} 
\definecolor{lightpurple}{RGB}{230,220,255}
\definecolor{lightgreen}{RGB}{210, 240, 210} 
\usepackage{CJKutf8}
\usepackage{booktabs}
\usepackage[table]{xcolor}
\usepackage{multirow} 
\usepackage{array}    

\def\confName{CVPR}
\def\confYear{2026}

\title{RVLF: A Reinforcing Vision–Language Framework for Gloss-Free Sign Language Translation}
\author{
Zhi Rao$^{1}$\quad
Yucheng Zhou$^{2}$\quad
Benjia Zhou$^{3}$\quad
Yiqing Huang$^{1}$\quad
Sergio Escalera$^{4}$\quad
Jun Wan$^{1}$\thanks{Corresponding author}
\\[4pt]
$^{1}$Macau University of Science and Technology  $^{2}$SKL-IOTSC, CIS, University of Macau \\
$^{3}$Beijing Institute of Technology, Zhuhai $^{4}$University of Barcelona\\
{\tt\small zhir@student.must.edu.mo,\;
yucheng.zhou@connect.um.edu.mo,\;
jwan@must.edu.mo}
}
\begin{document}
\maketitle
\begin{abstract}
Gloss-free sign language translation (SLT) is hindered by two key challenges: \textbf{inadequate sign representation} that fails to capture nuanced visual cues, and \textbf{sentence-level semantic misalignment} in current LLM-based methods, which limits translation quality.
To address these issues, we propose a three-stage \textbf{r}einforcing \textbf{v}ision–\textbf{l}anguage \textbf{f}ramework (\textbf{RVLF}). We build a large vision–language model (LVLM) specifically designed for sign language, and then combine it with reinforcement learning (RL) to adaptively enhance translation performance.
First, for a sufficient representation of sign language, RVLF introduces an effective semantic representation learning mechanism that fuses skeleton-based motion cues with semantically rich visual features extracted via DINOv2, followed by instruction tuning to obtain a strong SLT-SFT baseline. Then, to improve sentence-level semantic misalignment, we introduce a GRPO-based optimization strategy that fine-tunes the SLT-SFT model with a reward function combining translation fidelity (BLEU) and sentence completeness (ROUGE), yielding the optimized model termed SLT-GRPO. 
Our conceptually simple framework yields substantial gains under the gloss-free SLT setting without pre-training on any external large-scale sign language datasets, improving BLEU-4 scores by +5.1, +1.11, +1.4, and +1.61 on the CSL-Daily, PHOENIX-2014T, How2Sign, and OpenASL datasets, respectively.
To the best of our knowledge, this is the first work to incorporate GRPO into SLT. Extensive experiments and ablation studies validate the effectiveness of GRPO-based optimization in enhancing both translation quality and semantic consistency.
\end{abstract}

\section{Introduction}
\label{fig:intro}
Sign language translation (SLT) aims to bridge the communication gap between the deaf and hearing communities by translating sign/gesture videos \cite{gesture3, gesture1, gesture2} into spoken language text ~\citep{p14t,GFSLT-VLP}. Traditional gloss-based methods decompose SLT into two stages: sign-to-gloss and gloss-to-text ~\citep{SLRT, TS-SLT, SignBT, MMTLB}. Although glosses provide linguistic structure, they are inherently limited by high annotation costs, cross-linguistic inconsistencies, and annotation noise, hindering scalability for real-world deployment ~\citep{SLRT}.
With the rapid advancement of LLMs and VLM pre-training, gloss-free SLT has gained increasing attention as a more scalable and linguistically flexible alternative to gloss-based systems ~\citep{GFSLT-VLP,SIGN2GPT,Fla-LLM,Sign-LLM,SIGN2GPT,VAP,C2RL,Uni-Sign}. These approaches aim to directly map sign videos to spoken text, eliminating the dependency on costly gloss annotations. However, despite the expressive power of LLMs, current gloss-free methods still struggle to achieve accurate and fluent translations. 
We identify two core obstacles underlying this limitation. 
The first concerns the \textbf{inadequate sign representation}.  Existing models~\citep{GASLT, GFSLT-VLP, Fla-LLM, Sign-LLM, SIGN2GPT, C2RL, VAP, Uni-Sign} often rely on global image and/or skeleton-level features~\footnote{Uni-Sign introduces hand RGB images only as compensation when skeleton confidence is low, but its primary input remains skeletons.}, as shown in Fig.~\ref{fig:intro}~(a). However, accurate sign semantics is based on multiple cues, facial expressions, lip movements, hand gestures, and motion trajectories \cite{multistream-llm, face_sl, face_SLT, STMC, RTG-Net, signmusketeers}. Yet most gloss-free SLT models lack representations tailored to these characteristics. Without considering such cues, the extracted features fail to capture sufficient sign-specific semantics. 
The second obstacle lies in the \textbf{sentence-level semantic misalignment}. 
Previous models~\citep{GASLT, GFSLT-VLP, Fla-LLM, Sign-LLM, SIGN2GPT, C2RL, VAP} typically enhance sign representations through vision–language pre-training, and subsequently fine-tune an LLM using a cross-entropy (CE) objective that optimizes token-level likelihoods, as shown in Fig.~\ref{fig:intro} (a). 
\begin{figure}[!t]
\centering
\includegraphics[width=\linewidth]{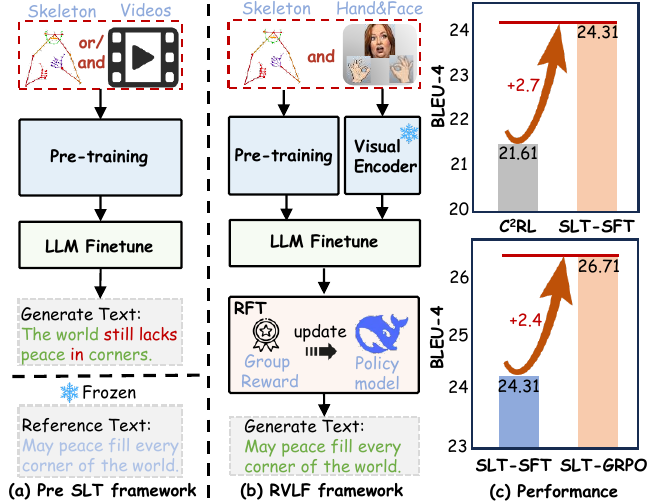}
\vspace{-5mm}
\caption{\small The overview illustrates the difference between previous gloss-free SLT frameworks (a) and the proposed RVLF framework (b), along with the performance of our method on the CSL-Daily \cite{SignBT} dataset (c). \emph{REF}: Reinforcement Fine-Tuning.}
\label{fig:method}
\vspace{-3mm}
\end{figure}
However, such token-wise optimization merely enforces local lexical accuracy without directly constraining the fidelity or completeness of the generated sentences. 
As a result, the model lacks explicit guidance to produce translations that are semantically aligned and coherent with the reference text.
To overcome the first limitation, an effective sign semantic representation is developed. Unlike relevant works~\cite{STMC, RTG-Net, signmusketeers, Uni-Sign}, the proposed approach performs cross-lingual visual–language pre-training on the skeleton encoder, using text as supervision to learn high-quality global representations of sign language cues, including motion trajectories, hand gestures, and lip movements. The resulting global representations are further fused with fine-grained hand and facial features extracted by DINOv2~\cite{dinov2}, enabling complementary modeling of facial expressions and enhanced representation of hand gestures, as shown in Fig.~\ref{fig:intro} (b). 
Following the standard LVLMs training paradigm~\cite{llava, llavanext, videollama2, videollama3}, we introduce a lightweight projector to map the fused sign visual features into the text embedding space of the LLM and conduct instruction tuning with LoRA to adapt the LLM for generating translation text. In this way, we achieve an SFT paradigm specifically tailored to the characteristics of sign language and obtain strong translation performance. We name this model SLT-SFT. As shown in Fig.~\ref{fig:intro} (c), our SLT-SFT improves BLEU-4 by 2.7 compared to the current State-of-the-Art (SOTA) model C$^2$RL \cite{C2RL} under the setting without pre-training on any external large-scale sign language datasets.
To address the second limitation, we introduce Group Relative Policy Optimization (GRPO) \cite{grpo} for the first time in the SLT domain and use it as a sentence-level reinforcement optimization strategy, forming our SLT-GRPO model as shown in Fig.~\ref{fig:intro} (b). 
Unlike standard supervised fine-tuning, reinforcement optimization enables the model to learn from its own generated translations by comparing them with reference sentences to compute sentence-level rewards that guide subsequent policy updates. In this framework, the translation process is formulated as a policy generation procedure, where rewards are derived from sentence-level metrics that evaluate the semantic and structural alignment between generated and reference sentences.
These metrics capture N-gram overlap and semantic coverage, encouraging the model to generate semantically equivalent outputs rather than merely high-probability tokens. Meanwhile, GRPO computes relative rewards within each group to stabilize optimization, reduce variance, and improve training efficiency. Overall, this method aligns the training objective with the evaluation metrics, leading to significant improvements in semantic consistency and translation accuracy. As shown in Fig.~\ref{fig:intro} (c), SLT-GRPO improves BLEU-4 by 2.4 compared to the SLT-SFT model.

Our main contributions are summarized as follows:

\begin{itemize}
\item We propose a three-stage RVLF:
(1) Pre-training: visual–language pre-training using skeleton-based input;
(2) SFT: fine-tuning the LLM using sufficient sign representations;
(3) RFT: sentence-level optimization using the GRPO strategy.

\item We introduce an effective and semantically expressive representation of sign language and fine-tune an LLM with strong linguistic capabilities to achieve high-quality sign language translation, referred to as SLT-SFT.
\item To address the sentence-level semantic misalignment in SLT, we apply GRPO-based reinforcement optimization with sentence-level rewards, forming SLT-GRPO.
\item RVLF achieves state-of-the-art results on multiple datasets and even surpasses Uni-Sign pre-trained on a large-scale external sign language dataset on CSL-Daily.
\end{itemize}
\section{Related Work}
\subsection{Sign Language Translation}
Sign Language Translation (SLT) aims to convert sign videos into spoken language sentences, posing unique challenges due to the spatiotemporal complexity of gestures and syntactic divergence from spoken language. Existing methods are broadly categorized into gloss-based and gloss-free approaches. Gloss-based methods leverage intermediate gloss annotations that provide structured, monotonic alignments between visual input and linguistic units, often incorporating auxiliary Sign Language Recognition (SLR) modules or directly decoding gloss sequences~\cite{SLRT, SignBT, TS-SLT, MMTLB}. However, gloss annotations are costly to obtain and limit scalability. To overcome this, gloss-free SLT seeks to directly map sign videos to spoken language without gloss supervision. Early studies~\cite{NSLT,SLRT,TSPNet,CSGCR,GASLT} demonstrated its feasibility, though a performance gap remains relative to gloss-based systems. Recent efforts employ visual-language pre-training to enhance representation learning. GFSLT~\cite{GFSLT-VLP}, for example, adapts CLIP~\cite{CLIP} for sign language, inspiring follow-up work~\cite{SIGN2GPT,Fla-LLM,Sign-LLM}. Other models~\cite{VAP,C2RL} leverage fine-grained alignment between sign segments and text, achieving competitive results. In parallel, large-scale pre-training on extensive sign corpora~\cite{YT-ASL-SLT,SSVP-SLT,Uni-Sign} has shown promise but comes at high computational and annotation costs. In contrast, our work focuses on exploring more comprehensive and semantically expressive representations of sign language at the algorithmic level, and leverages large language models with strong linguistic understanding and generation capabilities to decode such representations, thereby advancing the progress of gloss-free SLT.

\subsection{Reasoning Large Vision-Language Models}
Recent advances in Large Vision–Language Models (LVLMs) have greatly improved cross-modal understanding and generation. Early frameworks such as LLaVA~\cite{llava} introduced multimodal instruction tuning for large language models, substantially enhancing visual reasoning and alignment. The Qwen2.5-VL~\cite{Qwen2.5-VL} and Qwen3-VL~\cite{Qwen3-VL} series further expanded these capabilities by incorporating dynamic resolution and temporal modeling, achieving stronger multilingual comprehension and fine-grained video understanding. Recently, reinforcement learning (RL) has been shown to further enhance reasoning in large models. OpenAI o1~\cite{o1} demonstrated that outcome-based RL significantly improves the reasoning ability of LLMs, while DeepSeek-R1~\cite{deepseekR1} adopted rule-based rewards with the GRPO algorithm to teach structured reasoning. Building upon this, several works such as R1-OneVision~\cite{R1-OneVision}, VisualThinker-R1-Zero~\cite{VisualThinker-R1-Zero}, and VisionR1~\cite{VisionR1} have extended the R1 paradigm to multimodal settings, constructing step-by-step reasoning datasets and refining reinforcement fine-tuning strategies.
Inspired by these reasoning-enhanced LVLM paradigms, we introduce the RVLF for sign language translation, which integrates instruction tuning and GRPO-based reinforcement optimization to enhance semantic alignment.
\section{Method}\label{sec:method}
\begin{figure*}[!t]
\centering
\includegraphics[width=\linewidth]{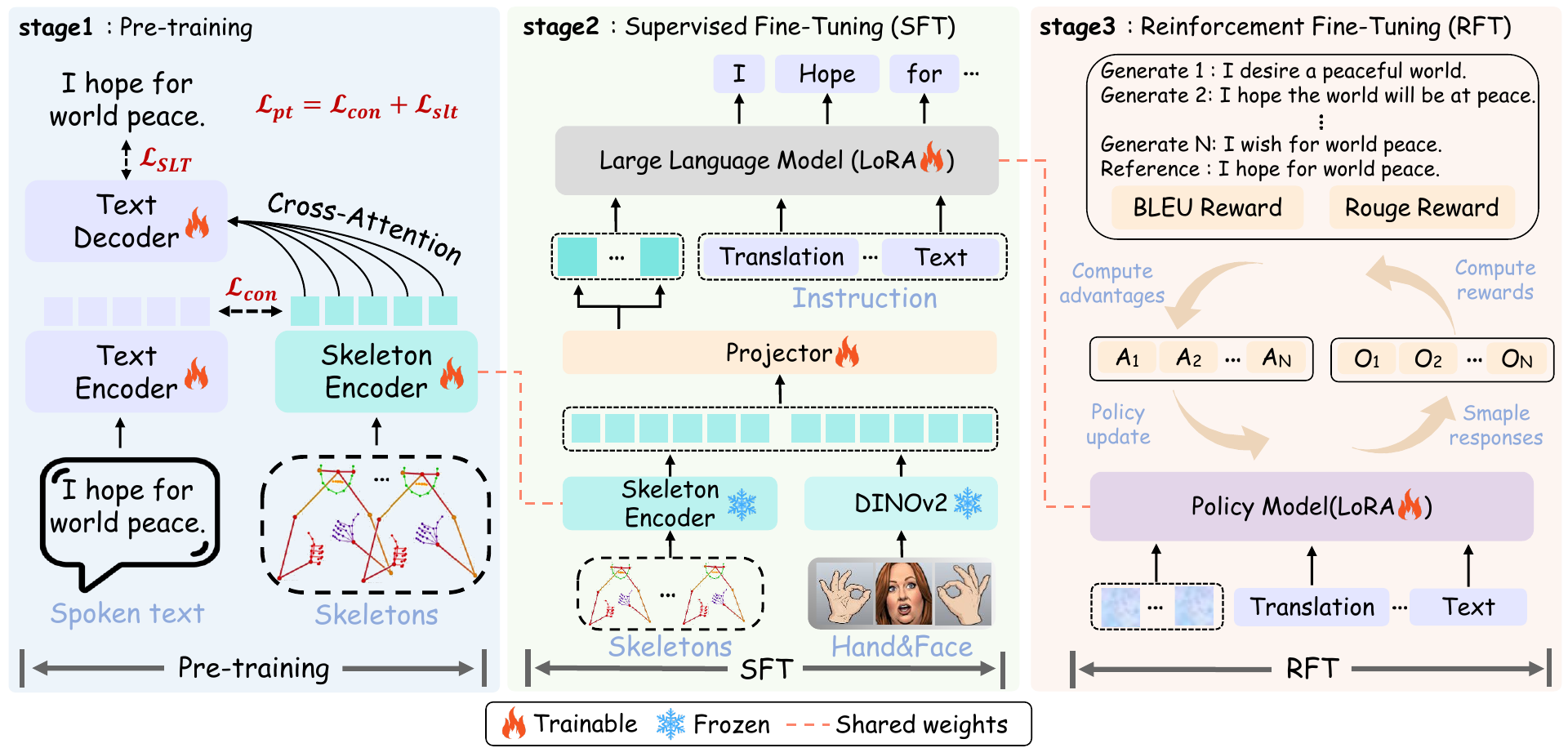}
\vspace{-3mm}
\caption{\small
    The framework of RVLF.
    \textbf{Stage 1 (Pre-training):} A vision-language foundation is built using contrastive ($\mathcal{L}_{con}$) and translation ($\mathcal{L}_{slt}$) losses.
    \textbf{Stage 2 (Supervised Fine-Tuning):} Adequate sign language representations are used as visual inputs, while appropriate instruction tuning is applied to the large language model to construct a vision–language model specifically tailored for sign language understanding and translation.
    \textbf{Stage 3 (Reinforcement Fine-Tuning):} A GRPO-based optimization strategy is employed to further fine-tune the policy model from the previous stage using a reward function based on sentence-level translation metrics, improving the overall translation performance.
}
\label{fig:SLT-R1}
\vspace{-1mm}
\end{figure*}

In this paper, we propose a three-stage training framework for sign language translation (Fig.~\ref{fig:SLT-R1}). \textbf{Stage 1} (Sec.~\ref{sec:pre-training}): We perform visual-language pre-training, aligning skeleton features with textual representations through combined contrastive and translation objectives. \textbf{Stage 2} (Sec.~\ref{sec:SFT}): We integrate the pre-trained skeleton features with semantically enriched hand–face features extracted by DINOv2~\cite{dinov2}, and fine-tune an LLM via instruction tuning to generate translations. \textbf{Stage 3} (Sec.~\ref{sec:RFT}): We apply GRPO optimization~\cite{grpo}, using sentence-level metrics as rewards to further refine translation quality.

\subsection{Vision-Language Pre-training}\label{sec:pre-training}
The goal of this stage is to learn robust multimodal representations by aligning sign language videos with their corresponding textual translations. Our pre-training framework is driven by two core objectives: a contrastive alignment objective and a translation objective.

\vspace{-4mm}
\paragraph{Contrastive Alignment.}
To establish a fine-grained alignment between visual and textual modalities, we introduce a contrastive learning objective. This requires encoding sign videos and text into comparable semantic spaces.

\noindent\textit{Skeleton Encoder.}
To capture the intricate dynamics of sign language, the Skeleton Encoder (\texttt{SkeletonEnc}) transforms a given sign video $V$ into a sequence of visual representations. First, we employ MMPose~\cite{mmpose} to extract 2D keypoints for the upper body, face, and hands from each frame, converting the video $V$ into a skeleton sequence $S$. This sequence is then processed by CoSign~\cite{cosign}, a GCN-based model that handles different sign groups (body, hands, face) independently to learn their unique co-occurrence patterns. The encoder thus yields a sequence of skeleton features:
\begin{align}
    Z_s = \text{SkeletonEnc}(V) = \{z_{s,1}, \dots, z_{s,m}\},
\end{align}
where $m$ is the number of frames in the video.

\noindent\textit{Text Encoder.}
The corresponding text $T$ is processed by the Text Encoder (\texttt{TextEnc}), which is derived from the initial layers of the mBART~\cite{mbart} encoder. This produces a sequence of semantic text embeddings:
\begin{align}
    Z_t = \text{TextEnc}(T) = \{z_{t,1}, \cdots, z_{t,l}\},
\end{align}
where $l$ is the number of tokens in the text.

\noindent\textit{InfoNCE Loss.}
Given a mini-batch of $B$ video-text pairs, the skeleton features $Z_s$ and text features $Z_t$ are padded to lengths $M$ and $L$, respectively. Both sequences are then projected into a $D$-dimensional embedding space by separate MLPs, resulting in $F_S \in \mathbb{R}^{M \times D}$ and $F_T \in \mathbb{R}^{L \times D}$. Let $f_i$ and $t_j$ be the L2-normalized row vectors of $F_S$ and $F_T$. The cross-modal similarity matrix $\bm{E} \in \mathbb{R}^{M \times L}$ is then computed as:
\begin{align}
    E_{i,j} = f_i \cdot t_j^\top.
\end{align}

To calculate the global sign-skeleton-to-text similarity $z_{S,T}$, a row-wise Softmax~\cite{cico} is applied to $\bm{E}$ to obtain the attention weights $\bm{P}$:
\begin{align}
    P_{i,j} = \frac{\exp(E_{i,j} / \tau)}{\sum_{k=1}^L \exp(E_{i,k} / \tau)},
\end{align}
where $\tau$ denotes a trainable temperature parameter. A re-weighted similarity is then computed for each skeleton segment:
\begin{align}
    e_{S \rightarrow T, i} = \sum_{j=1}^{L} (P_{i,j} \times E_{i,j}).
\end{align}
Finally, the global similarity is computed by averaging over all skeleton segments:
\begin{align}
    z_{S,T} = \frac{1}{M} \sum_{i=1}^{M} e_{S \rightarrow T, i}.
\end{align}

Symmetrically, the global text-to-sign-skeleton similarity $z_{T,S}$ is computed. To apply contrastive learning across the batch, we compute these global similarity scores for every skeleton sequence $S^{(b)}$ and every text sequence $T^{(b')}$ in the mini-batch (where $b, b' \in \{1, \dots, B\}$). This populates two matrices, $\bm{Z}_{S2T}, \bm{Z}_{T2S} \in \mathbb{R}^{B \times B}$, where the element $\bm{Z}_{S2T}^{(b, b')} = z_{S^{(b)}, T^{(b')}}$ represents the similarity from the $b$-th skeleton to the $b'$-th text. A bidirectional InfoNCE loss is then applied to these matrices:
\begin{align}
    \mathcal{L}_{S2T} &= -\frac{1}{B} \sum_{b=1}^B \log \frac{\exp(Z_{S2T}^{(b,b)} / \tau')}{\sum_{b'=1}^B \exp(Z_{S2T}^{(b,b')} / \tau')}, \\
    \mathcal{L}_{T2S} &= -\frac{1}{B} \sum_{b=1}^B \log \frac{\exp(Z_{T2S}^{(b,b)} / \tau')}{\sum_{b'=1}^B \exp(Z_{T2S}^{(b,b')} / \tau')}, \\
    \mathcal{L}_{\text{con}} &= \beta \mathcal{L}_{S2T} + (1 - \beta)\mathcal{L}_{T2S},
\end{align}
where $\tau'$ is a temperature parameter and $\beta$ balances the two directional losses.

\vspace{-4mm}
\paragraph{Translation Objective.}
To equip the model with generative capabilities, we employ a translation objective. The skeleton representations $Z_s$ are fed into a Text Decoder, which is adapted from the mBART decoder, to generate a textual translation. This is optimized via a standard cross-entropy loss, similar to a captioning task~\cite{coca}:
\begin{equation}
\mathcal{L}_{\text{slt}} = - \sum_{i=1}^{L} \log P(y_i \mid y_{<i}, Z_s),
\end{equation}
where $y_i$ is the $i$-th token in the ground-truth sentence and $L$ is the padded sequence length.

\vspace{-4mm}
\paragraph{Pre-training Objective.}
The final pre-training objective combines the contrastive and translation losses to learn representations that are both well-aligned and generative:
\begin{equation}
\mathcal{L}_{\text{pt}} = \mathcal{L}_{\text{con}} + \mathcal{L}_{\text{slt}}.
\end{equation}

\subsection{Supervised Fine-Tuning}\label{sec:SFT}
In the second stage, we fine-tune the LLM using the learned sign representations. In addition to the skeleton embeddings, we incorporate dense visual cues extracted by DINOv2~\cite{dinov2}. It is motivated by the observation that the sparse skeletal modality alone is insufficient to capture the full expressiveness of sign language, as crucial semantic cues often lie in fine-grained hand shapes and facial expressions. 

\vspace{-4mm}
\paragraph{Dense Visual Cue Extraction.} To capture fine-grained motion and expression details, we extract dense visual cues from hand and face regions. Specifically, we first detect bounding boxes for the hands and face in each video frame, then crop these regions and feed them into a pre-trained DINOv2~\cite{dinov2} model to obtain semantically rich feature sequences. To align feature dimensions for fusion, the extracted features are linearly projected to match the skeleton representation. The fused enhanced representation is computed as:
\begin{equation}
    Z_e = Z_s + \alpha \cdot Z_{face} + \beta \cdot Z_{hand}
\end{equation}
where $\alpha$ and $\beta$ are tunable coefficients controlling the relative contributions of facial and hand cues.
Finally, the enhanced embedding is projected into the target latent space via a learnable MLP transformation:
\begin{equation}
Z = \mathrm{MLP}(Z_{e}).
\end{equation}

\vspace{-4mm}
\paragraph{LLM Fine-Tuning.}  We employ Qwen3~\cite{qwen3} model as the base large language model (LLM) for sign language translation. The aligned sign features $\mathbf{Z}$ are prepended to the LLM’s input embeddings. The model is then fine-tuned using LoRA-based instruction tuning to autoregressively generate the target textual sequence, written as:
\begin{equation}
\mathcal{L}_{\mathrm{sft}} = -\sum_{t=1}^{L} \log p_\theta(y_t \mid y_{<i}, Z),
\end{equation}
where $\theta$ denotes the trainable LoRA parameters in the LLM. 

\subsection{Reinforcement Fine-Tuning}\label{sec:RFT}
SFT optimizes models at the token level via cross-entropy, thereby overlooking sentence-level qualities such as semantic fidelity and completenes. To align training with these holistic objectives, a reinforcement fine-tuning stage using the GRPO is introduced.

The GRPO algorithm begins by sampling $N$ candidate outputs $\{O_1, \ldots, O_N\}$ from the current policy $\pi_{\theta}$ for a given query prompt $q$. Each response $O_i$ is then evaluated by a reward function $R(q, O_i)$ to obtain a raw reward score $r_i$. 
In our SLT task, the prompt $q$ corresponds to the fused sign features $Z_{sign}$, and the candidate outputs $\{O_1, \ldots, O_N\}$ are the translations generated by the policy. The reward function $R(q, O_i)$ is designed to measure the quality of a generated translation $O_i$ against the ground-truth reference sentence $T$. It is computed as a combination of BLEU-4 and ROUGE-L scores, which capture N-gram overlap and semantic coverage, respectively:
\begin{equation}
    \!\!R(q, O_i) \!=\! \lambda \!\cdot\! \text{BLEU-4}(O_i, T) \!+\! (1\!-\!\lambda) \!\cdot\! \text{ROUGE}(O_i, T)
\end{equation}
This reward directly incentivizes the model to generate outputs that are semantically and structurally superior at the sentence level.

To measure the relative quality of each response within the sampled group, GRPO standardizes the raw rewards to obtain the advantage value in Eq.~\ref{eq:advantage}. The advantage value ${A}_i$ denotes the normalized advantage of the response $O_i$ relative to other samples within the group.

\begin{equation}
    A_i = \frac{r_i - \text{mean}\{r_1, r_2, \ldots, r_N\}}{\text{std}\{r_1, r_2, \ldots, r_N\}}
    \label{eq:advantage}
\end{equation}

The policy $\pi_{\theta}$ is updated with a training objective (Eq.~\ref{eq:grpo_main}), designed to encourage the generation of responses with higher advantages.
\begin{align}
    &\mathcal{J}_{\text{GRPO}}(\theta) = 
    \mathbb{E}_{\{O_i\}_{i=1}^{N} \sim \pi_{\text{old}}(\cdot | q)}
    \left[
        \frac{1}{N} \sum_{i=1}^{N}
        L^{(i)}_{\text{GRPO}}
    \right]
    \label{eq:grpo_main}
    \\
    &L^{(i)}_{\text{GRPO}} = 
    \min \left( c_1 \cdot A_i,\; c_2 \cdot A_i \right)
    - \beta D_{\text{KL}}(\pi_{\theta} \| \pi_{\text{ref}})
    \label{eq:grpo_inner}
    \\
    &\!\!\!c_1 \!=\! \frac{\pi_{\theta}(O_i \mid q)}{\pi_{\text{old}}(O_i \mid q)},~\!
    c_2 \!=\! \text{clip}\left(
        \frac{\pi_{\theta}(O_i \mid q)}{\pi_{\text{old}}(O_i \mid q)},
        1 \!-\! \varepsilon,
        1 \!+\! \varepsilon
    \!\right)\!\!
    \label{eq:clip}
\end{align}
where $D_{\text{KL}}(\pi_{\theta} \| \pi_{\text{ref}})$ denotes the KL divergence between the current policy $\pi_{\theta}$ and the reference policy $\pi_{\text{ref}}$, which serves as a regularization term to prevent large deviations. The clipping mechanism in $c_2$ stabilizes training by constraining the policy update ratio.

\section{Experiment}

\subsection{Experimental Setup}

\paragraph{Datasets.} We evaluated our method on four mainstream sign language datasets: CSL-Daily\cite{SignBT}, PHOENIX-2014T\cite{p14t}, How2Sign\cite{how2sign} and OpenASL\cite{OpenASL}. PHOENIX-2014T is a benchmark dataset for German Sign Language to spoken German translation. It is collected from German weather broadcast programs. The CSL-Daily dataset is a Chinese sign language (CSL) dataset recorded in the laboratory whose topic focuses on daily life. How2Sign is a large-scale American Sign Language (ASL) dataset comprising approximately 80 hours of sign language videos. Collected from instructional videos across diverse categories, the dataset represents a more open-domain setting. The OpenASL dataset is an ASL dataset collected from online videos that cover various topics.
\paragraph{Evaluation Metrics.}
We evaluate translation performance using standard metrics commonly employed in sign language translation, including BLEU and ROUGE-L scores. 
BLEU evaluates \textbf{translation fidelity} by measuring the overlap of n-grams between the model output and the reference translation~\cite{bleu}, reflecting how well the generated text preserves the original meaning. In contrast, ROUGE-L assesses \textbf{sentence completeness} by calculating the length of the longest common subsequence between the generated and reference texts~\cite{rouge}, indicating the overall structural and semantic coverage of the translation.

\begin{table}[t]\small
  \centering
  \resizebox{\linewidth}{!}{
  \begin{tabular}{lccccccc}
  \toprule
    \textbf{Method} & \textbf{ROUGE} & \textbf{B@1} & \textbf{B@2} & \textbf{B@3} & \textbf{B@4} \\
    \midrule
     \rowcolor{gray!15}\multicolumn{6}{c}{\bf \textit{Gloss-based}} \\
     \midrule
     SLRT$^\dag$~\cite{SLRT} & 36.74 & 37.38 & 24.36 & 16.55 & 11.79 \\
     SignBT~\cite{SignBT}    & 49.31 & 51.42 & 37.26 & 27.76 & 21.34 \\
     MMTLB~\cite{MMTLB}      & 53.25 & 53.31 & 40.41 & 30.87 & 23.92 \\
     SLTUNET~\cite{SLTUNET}  & 54.08 & 54.98 & 41.44 & 31.84 & 25.01 \\
     TS-SLT~\cite{TS-SLT}    &\underline{55.72} &\underline{55.44}&\underline{42.59} & \underline{32.87} &\underline{25.79} \\
     CV-SLT~\cite{CV-SLT}    &\textbf{57.06} &  \textbf{58.29} & \textbf{45.15} & \textbf{35.77} & \textbf{28.94} \\
     \midrule
     \rowcolor{gray!15}\multicolumn{6}{c}{\bf \textit{Gloss-free}} \\
     \midrule
    \textcolor{gray}{Uni-Sign~\cite{Uni-Sign}} & \textcolor{gray}{56.51} & \textcolor{gray}{55.08} & \textcolor{gray}{42.14} & \textcolor{gray}{32.98} & \textcolor{gray}{26.36} \\
     \midrule
     TSPNet$^*$~\cite{TSPNet}    & 18.38 & 17.09  & 8.98 & 5.07 & 2.97 \\
     GASLT~\cite{GASLT}          & 20.35  & 19.90  & 9.94& 5.98 & 4.07  \\
     NSLT$^\dag$~\cite{NSLT}     & 34.54 & 34.16 & 19.57 & 11.84 & 7.56 \\
     GFSLT-VLP~\cite{GFSLT-VLP}  & 36.44 & 39.37 & 24.93 & 16.26 & 11.00 \\
     {Sign2GPT}~\cite{SIGN2GPT}  & {42.36}& {41.75} &{28.73} &{20.60}& {15.40}  \\
     {SignLLM}~\cite{Sign-LLM}   & {39.91} & {39.55} &{28.13}& {20.07} &{15.75}\\
     {VAP}~\cite{VAP}            & {48.56}  & {49.99} & - & - & {20.85} \\
     {C$^2$RL}~\cite{C2RL}       & 48.21    & 49.32 & 36.28 & 27.54 & 21.61\\
     SLT-SFT(Ours)  &\underline{52.43} &\underline{52.34} & \underline{39.03} &\underline{30.32} &\underline{24.31}\\
     SLT-GRPO(Ours)  &\textbf{55.92} & \textbf{54.96} & \textbf{42.45} & \textbf{33.33} & \textbf{26.71}\\
    \bottomrule
    \end{tabular}}
    \caption{SLT results on CSL-Daily dataset. 
    $*$ denotes methods reproduced by \protect\cite{GASLT}. $\dag$ denotes methods reproduced by \protect\cite{SignBT}.
    The result of the work trained on large-scale external sign language dataset (CSL-News \cite{Uni-Sign}) is shown in \textcolor{gray}{gray}.
    The best scores are \textbf{bolded}. The second-best scores are \underline{underlined}.
   }
    \label{tab:CSL-Daily}
    \vspace{-2mm}
\end{table}

\paragraph{Implementation details.}
We use CoSign~\cite{cosign} and a three-layer Transformer encoder as our skeleton encoder. The text encoder and decoder use the fully transformer blocks, which consist of three layers each in both the encoder and decoder. Each layer includes 8 attention heads, a hidden size of 512, and a feed-forward dimension of 2048. The transformer encoder of skeleton encoder shares the same configuration as the text encoder but utilizes only the encoder component. Unlike previous work \cite{llava}, we conduct experiments on four different sign language datasets in Chinese, German, and American, using a unified decoder-only large language model (Qwen3). All experiments are conducted on four NVIDIA V100 GPUs (32GB) in float16 precision  to optimize memory efficiency. During pre-training, we employ the SGD \cite {SGD} optimizer with a learning rate of 0.01 and a weight decay of 0.001. During supervised fine-tuning, we employed the AdamW optimizer \cite{AdamW}. The LLM was trained with a learning rate of $2 \times 10 ^{-4}$, while the multimodal projector used a learning rate of $2 \times 10 ^{-5}$. The weight decay was set to 0. During reinforcement fine-tuning, we employed the AdamW optimizer \cite{AdamW}. Trained with a learning rate of $2 \times 10 ^{-5}$. The three stages are trained for 200, 40, and 2 epochs, respectively. For each stage, the model achieving the highest BLEU-4 score on the validation set is selected as the final model. $\alpha$, $\beta$, and $\lambda$ are set as 1, 1, and 0.5, respectively. More implementation details are provided in the supplementary material.
\begin{table}[!t]\small
    \centering
      \resizebox{\linewidth}{!}{
      \begin{tabular}{lccccc}
        \toprule
        \textbf{Method} & \textbf{ROUGE} & \textbf{B@1} & \textbf{B@2} & \textbf{B@3} & \textbf{B@4} \\
        \midrule
        \rowcolor{gray!15}\multicolumn{6}{c}{\bf \textit{Gloss-based}} \\
        \midrule
        SLRT~\cite{SLRT}        & -     & 46.61 & 33.73 & 26.19 & 21.32 \\
        SignBT~\cite{SignBT}    & 49.54 & 50.80 & 37.75 & 29.72 & 24.32 \\
        TS-SLT~\cite{TS-SLT}    &\underline{53.48} &\textbf{54.90} & \underline{42.43} &\underline{34.46} & \underline{28.95} \\
        CV-SLT~\cite{CV-SLT}    & \textbf{54.33} & \underline{54.88} & \textbf{42.68} & \textbf{34.79} & \textbf{29.27} \\
        \midrule
        \rowcolor{gray!15}\multicolumn{6}{c}{\bf \textit{Gloss-free}} \\
        \midrule
        NSLT~\cite{NSLT}      & 31.80 & 32.24 & 19.03 & 12.83 & 9.58  \\
        SLRT-GF*~\cite{SLRT}  & 31.10 & 30.88 & 18.57 & 13.12 & 10.19 \\
        TSPNet~\cite{TSPNet}    & 34.96 & 36.10 & 23.12 & 16.88 & 13.41 \\
        CSGCR~\cite{CSGCR}     & 38.85 & 36.71 & 25.40 & 18.86 & 15.18 \\
        GASLT~\cite{GASLT}     & 39.86 & 39.07 & 26.74 & 21.86 & 15.74 \\
        GFSLT-VLP~\cite{GFSLT-VLP} & 42.49 & 43.71 & 33.18 & 26.11 & 21.44 \\
        Sign2GPT~\cite{SIGN2GPT}  & 48.90 & 49.54 & 35.96 & 28.83 & 22.52 \\
        FLa-LLM~\cite{Fla-LLM}   & 45.27 & 46.29 & 35.33 & 28.03 & 23.09 \\
        Sign-llm~\cite{Sign-LLM}  & 44.49 & 45.21 & 34.78 & 28.05 & 23.40 \\
        LLaVA-SLT~\cite{LLaVA-SLT} & 50.44 & 51.20 & 37.51 & 29.39 & 23.42 \\
        VAP~\cite{VAP} & 51.28 & 53.07 & -- & -- & 26.16 \\
        C$^2$RL~\cite{C2RL} & 50.96 & 52.81 & 40.20 & 32.20 & 26.75 \\
        SLT-SFT(Ours)   &\underline{51.81} &\underline{53.29} & \underline{40.44} &\underline{32.35} &\underline{26.83} \\
        SLT-GRPO(Ours)   & \textbf{52.56} & \textbf{53.60} & \textbf{41.20} & \textbf{33.24} & \textbf{27.86} \\
        \bottomrule
    \end{tabular}

    }
    \vspace{-2mm}
    \caption{Experimental results on PHOENIX14T dataset. 
    No prior work has been pre-trained on large-scale external sign language datasets and evaluated on the PHOENIX-2014T dataset.
    }
    \label{tab:P14T}
    \vspace{-4mm}
\end{table}

\subsection{Main Results}
\textbf{Experiment on  CSL-Daily Datasets.} 
Table \ref{tab:CSL-Daily} summarizes the comparison with previous SLT models. In the gloss-free setting for fair comparison, our SLT-SFT achieves strong gains over all methods that do not use large-scale external sign-language pre-training, outperforming the previous SOTA C$^2$RL \cite{C2RL}  by a clear margin (24.31 vs. 21.61 in BLEU-4). Despite this progress, a gap remains to Uni-Sign~\cite{Uni-Sign} (26.36), which benefits heavily from costly large-scale pre-training. Furthermore, our sentence-level optimization further enforces translation fidelity, enabling SLT-GRPO to boost BLEU-4 by +2.4 and surpass Uni-Sign (26.71 vs. 26.36). SLT-GRPO also substantially outperforms  C$^2$RL \cite{C2RL}  (+5.1). Moreover, SLT-GRPO shows competitive or superior performance to several gloss-based methods, outperforming SignBT (26.71 vs. 21.34) and TS-SLT (26.71 vs. 25.79), and approaching the gloss-based SOTA CV-SLT (26.71 vs. 28.94). These results collectively highlight the advantages and efficiency of our method across gloss-free and gloss-based settings.

\noindent\textbf{Experiment on PHOENIX14T Dataset.}
As shown in Table \ref{tab:P14T}, SLT-SFT slightly surpasses the previous SOTA C$^2$RL (26.83 vs. 26.75) in the gloss-free setting. And the propsoed sentence-level optimization further improves performance: SLT-GRPO reaches 27.86 BLEU-4, establishing a new gloss-free SOTA (+1.11 over C$^2$RL). Despite using no gloss supervision, SLT-GRPO outperforms early gloss-based models (SLRT 21.32; SignBT 24.32) and narrows the gap to the gloss-based SOTA CV-SLT (29.27). 

\begin{table}[!t]\small
  \centering
  \resizebox{\linewidth}{!}{
  \begin{tabular}{lccccc}
    \toprule
    \textbf{Method}     & \textbf{ROUGE}  & \textbf{B@1}   & \textbf{B@2}   & \textbf{B@3}   & \textbf{B@4} \\
    \midrule
    \textcolor{gray}{YouTube-ASL~\cite{YT-ASL-SLT}} & \textcolor{gray}{-} & \textcolor{gray}{37.8} & \textcolor{gray}{24.1} & \textcolor{gray}{16.9} & \textcolor{gray}{12.4} \\
    \textcolor{gray}{SignMusketeers~\cite{signmusketeers}} & \textcolor{gray}{-} & \textcolor{gray}{41.5} & \textcolor{gray}{27.2} & \textcolor{gray}{19.3} & \textcolor{gray}{14.3} \\
    \textcolor{gray}{Uni-Sign~\cite{Uni-Sign}} & \textcolor{gray}{36.0} & \textcolor{gray}{40.2} & \textcolor{gray}{27.1} & \textcolor{gray}{19.7} & \textcolor{gray}{14.9} \\
    \textcolor{gray}{SSVP-SLT~\cite{SSVP-SLT}} & \textcolor{gray}{38.4} & \textcolor{gray}{43.2} & \textcolor{gray}{28.8} & \textcolor{gray}{20.8} & \textcolor{gray}{15.5} \\
    \midrule
    YouTube-ASL~\cite{YT-ASL-SLT} & --    & 15.0  & 5.1   & 2.3   & 1.2  \\
    GloFE-VN~\cite{GloFE-VN}     & 12.6  & 14.9 & 7.3    & 3.9   & 2.2  \\
    SSVP-SLT~\cite{SSVP-SLT}     & 25.7  & 30.2 & 16.7   & 10.5  & 7.0  \\
    OpenSLT~\cite{OpenSLT}       & --    & 34.0 & 19.3   & 12.2  & 8.0  \\
    C$^2$RL~\cite{C2RL}          & 27.0  & 29.1 & 18.6 & 12.9      & 9.4  \\
    FLa-LLM~\cite{Fla-LLM}       & 27.8  & 29.8 & 19.0 & 13.3 & 9.7  \\
    VAP~\cite{VAP}               & 27.8  & \underline{39.2} & --    & --    & \underline{12.9}  \\
    SLT-SFT (Ours)               & \underline{31.3}  & 38.3 & \underline{23.7} & \underline{16.5} & 12.2 \\
    SLT-GRPO (Ours)              & \textbf{34.7}  & \textbf{39.4} & \textbf{25.3} & \textbf{18.6} & \textbf{14.3} \\
    \bottomrule
  \end{tabular}
  }
  \vspace{-2mm}
  \caption{Experimental results on the How2Sign dataset. 
  The results of models pre-trained on a large-scale external sign language dataset (Youtube-ASL \cite{YT-ASL-SLT}) are shown in \textcolor{gray}{gray}. 
  }
  \label{tab:How2Sign}
  \vspace{-2mm}
\end{table}

\noindent\textbf{Experiment on How2Sign Dataset.} As shown in Table \ref{tab:How2Sign} , SLT-SFT attains a slightly lower BLEU-4 than VAP (12.2 vs. 12.9) but achieves a higher ROUGE (31.3 vs. 27.8), indicating stronger semantic completeness. SLT-GRPO further lifts BLEU-4 to 14.3, establishing a new gloss-free SOTA (+1.4 over VAP). Despite using no large-scale sign-language pre-training, SLT-GRPO surpasses Youtube-ASL, performs comparably to SignMusketeers, and approaches Uni-Sign and SSVP-SLT (14.9 and 15.5).

\noindent\textbf{Experiment on OpenASL Dataset.} 
Table \ref{tab:OpenASL} reports results on OpenASL, which, like How2Sign, lacks gloss annotations; all comparisons are gloss-free. SLT-SFT slightly outperforms the previous SOTA VAP (21.28 vs. 21.23), and SLT-GRPO further improves performance to 22.84, setting a new SOTA (+1.61 over VAP). Despite the absence of large-scale external pre-training, SLT-GRPO remains competitive with Uni-Sign (22.84 vs. 23.14).

\begin{table}[!t]\small
  \centering
  \resizebox{\linewidth}{!}{
  \begin{tabular}{lccccc}
    \toprule
    \textbf{Method}                         & \textbf{ROUGE}  & \textbf{B@1}   & \textbf{B@2}   & \textbf{B@3}   & \textbf{B@4}   \\
    \midrule
    \textcolor{gray}{Uni-Sign~\cite{Uni-Sign}} & \textcolor{gray}{43.22} & \textcolor{gray}{49.35} & \textcolor{gray}{36.22} & \textcolor{gray}{28.55} & \textcolor{gray}{23.14} \\
    \midrule
     Conv-GRU$^\dag$~\cite{NSLT}            & 16.10 & 16.11 & 8.85  & 6.18  & 4.58\\
     I3D-transformer~\cite{OpenASL}         & 18.64 & 18.31 & 10.15 & 7.19  & 5.66\\
     OpenASL~\cite{OpenASL}                 & 21.02 & 20.92 & 12.08 & 8.59  & 6.72  \\
     GloFE-VN~\cite{GloFE-VN}               & 21.75 & 21.56 & 12.74 & 9.05  & 7.06 \\
     C$^2$RL ~\cite{C2RL}                   & 31.36 & 31.46 & 21.85 & 16.58 & 13.21 \\
     VAP~\cite{VAP}                         & 41.38 & 45.92 & -     & -     & 21.23 \\ 
    SLT-SFT (Ours)                          & \underline{41.84}  & \underline{46.03} & \underline{33.64} & \underline{26.31} & \underline{21.28} \\
    SLT-GRPO (Ours)    & \textbf{43.92} & \textbf{48.73} & \textbf{35.40} & \textbf{27.99} & \textbf{22.84} \\
    \bottomrule
  \end{tabular}
  }
  \vspace{-2mm}
  \caption{Experimental results on OpenASL dataset. 
  The result of the work trained on a large-scale external sign language dataset (Youtube-ASL \cite{YT-ASL-SLT}) is shown in \textcolor{gray}{gray}.
  }
  \label{tab:OpenASL}
  \vspace{-2mm}
\end{table}

\begin{table}[!t]\small
    \centering
    \setlength{\tabcolsep}{1.8mm} 
    \begin{tabular}{lcccccc}
        \toprule
        \textbf{Setting}  & \textbf{ROUGE} & \textbf{B@1} & \textbf{B@2} & \textbf{B@3} & \textbf{B@4} \\
        \midrule
        \rowcolor{cyan!10}SLT-GRPO  & 55.92 & 54.96 & 42.45 & 33.33 & 26.71 \\
        \midrule
        w/o Pre-training           & 42.10 & 41.68 & 29.68 & 22.20 & 17.22 \\
        w/o DINOv2                 & 51.45 & 49.51 & 37.66 & 28.10 & 22.93 \\
        w/o RFT                    & 53.81 & 52.34 & 39.03 & 30.32 & 24.31 \\
        \bottomrule
    \end{tabular}
  \vspace{-2mm}
    \caption{Comparison of evaluation metrics across different settings. 
    \emph{w/o Pre-training}: SLT-SFT model trained with skeleton features without pre-training;
    \emph{w/o DINOv2}: SLT-SFT model trained without hand or facial features extracted by DINOv2;
    \emph{w/o RFT}: Equivalent to the SLT-SFT model.
    }
    \label{tab:metrics_comparison}
    \vspace{-2mm}
\end{table}

\begin{table}[!t]\small
    \centering
    \begin{tabular}{cc|ccccc}
        \toprule
        \textbf{Face} & \textbf{Hand} & \textbf{ROUGE} & \textbf{B@1} & \textbf{B@2} & \textbf{B@3} & \textbf{B@4} \\
        \midrule
        $\times$     & $\times$     & 51.45 & 49.51 & 37.66 & 28.10 & 22.93 \\
        $\checkmark$ & $\times$     & 52.84  & 52.12  & 38.73  & 29.86  & 23.69 \\
        $\times$ & $\checkmark$     & 52.74  & 51.32  & 38.10  & 29.35  & 23.27 \\
        $\checkmark$ & $\checkmark$ & 53.81  & 52.34  & 39.03  & 30.32  & 24.31 \\
        \bottomrule
    \end{tabular}
  \vspace{-2mm}
    \caption{Comparison of SLT-SFT model performance under various DINOv2 features. $\checkmark$ represents the use of the corresponding input feature.}
    \label{tab:face_hand}
  \vspace{-2mm}
\end{table}

\subsection{Ablation Study.}
To further investigate the effectiveness of each component of the proposed RVLF method, we conduct extensive ablation experiments on the CSL-Daily dataset.

\vspace{-4mm}
\paragraph{Ablation on the framework.} 
Table~\ref{tab:metrics_comparison} presents an ablation study across different framework settings. Removing the pre-training stage results in a significant performance drop (-7.09 BLEU@4), highlighting the domain gap between sign language videos and general video data, and underscoring the need for a specialized skeleton encoder to learn high-quality representations of sign language cues. Text-supervised training effectively guides the encoder toward more discriminative semantic representations. Removing the fine-grained sign semantics from DINOv2 leads to a -1.38 BLEU@4 drop, demonstrating the value of hand and facial features in capturing sign-relevant information. These features contribute to complementary modeling of facial expressions and enhanced representation of hand gestures. Finally, removing the RFT stage causes a -2.40 BLEU@4 drop, indicating that sentence-level reinforcement optimization improves semantic consistency and translation accuracy by encouraging the model to consider sentence fidelity rather than just high-probability tokens.

\vspace{-4mm}
\paragraph{Ablation on DINOv2 features.} As shown in Table~\ref{tab:face_hand}, adding either hand or face features extracted by DINOv2 results in varying degrees of performance improvement, indicating that both types of features provide complementary semantic cues to the sign representations. Notably, incorporating face features yields larger gains, likely because facial expressions offer more informative cues for enriching the global representations learned by the skeleton encoder. Moreover, combining both hand and face features achieves even greater improvements than using either alone, highlighting the strong complementarity between these two forms of fine-grained sign semantics.

\begin{table}[ht]\small
    \centering
    \setlength{\tabcolsep}{4.0pt} 
    \begin{tabular}{cc|ccccc}
        \toprule
        \textbf{BLEU-4} & \textbf{ROUGE} & \textbf{ROUGE} & \textbf{B@1} & \textbf{B@2} & \textbf{B@3} & \textbf{B@4} \\
        \midrule
        $\checkmark$ & $\times$     & 55.26 & 55.26 & 41.97 & 32.78 & 26.23 \\
        $\times$ & $\checkmark$     & 55.03 & 54.17 & 41.48 & 32.20 & 25.66 \\
        $\checkmark$ & $\checkmark$ & 55.92 & 54.96 & 42.45 & 33.33 & 26.71 \\
        \bottomrule
    \end{tabular}
    \caption{Comparison of evaluation metrics under different reward configurations. $\checkmark$ denotes the use of the corresponding reward function.}
    \label{tab:reward_func}
\end{table}

\vspace{-4mm}
\paragraph{Ablation on the reward functions.} We conduct experiments using different reward functions. As shown in Table~\ref{tab:reward_func}, using either BLEU or ROUGE alone already yields competitive performance, while combining the two leads to further improvements. This indicates that, under the sentence-level reinforcement optimization strategy, jointly using BLEU and ROUGE as rewards effectively constrains both the fidelity and completeness of the generated sentences, ultimately resulting in better translation quality.

\subsection{Analysis}

\begin{figure}[!t]
    \centering
    \includegraphics[width=0.9\linewidth]{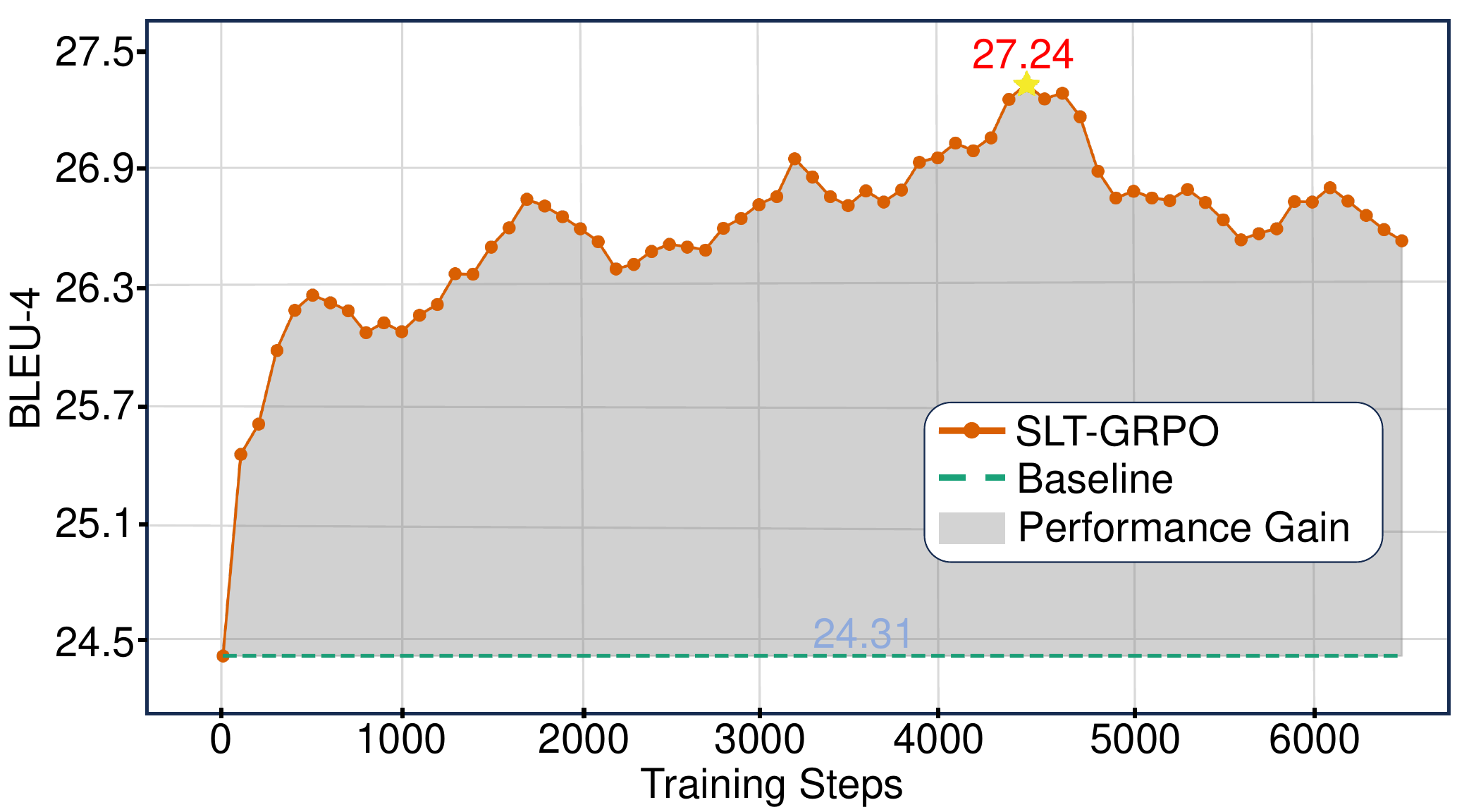}
    \vspace{-1mm}
    \caption{
    Learning curves of SLT-GRPO on the CSL-Daily dataset. The baseline corresponds to the SLT-SFT model. The performance shown in the SLT-GRPO curve reflects results on the validation set of CSL-Daily.
    }
    \label{fig:reward}
\end{figure}

\begin{CJK*}{UTF8}{gbsn}

\definecolor{mygreen}{RGB}{198,239,206}
\definecolor{myyellow}{RGB}{255,242,204}
\definecolor{mypurple}{RGB}{218,210,233}
\definecolor{myred}{RGB}{255,204,204}
\newcommand{\hlg}[1]{\colorbox{mygreen}{#1}}
\newcommand{\hly}[1]{\colorbox{myyellow}{#1}}
\newcommand{\hlp}[1]{\colorbox{mypurple}{#1}}
\newcommand{\hlr}[1]{\colorbox{myred}{#1}}

\begin{table}[!t]
\centering
\scriptsize
\renewcommand{\arraystretch}{1.25}
\setlength{\tabcolsep}{3pt}

\begin{tabular}{@{}p{0.12\linewidth}|p{0.68\linewidth}|p{0.15\linewidth}@{}}
\toprule
\textbf{Source} & \textbf{Sentence} & \textbf{Metrics} \\
\midrule

\textbf{Reference} &
世界上没有后悔药 \newline
\textit{(There is no medicine for regret in the world.)} &
~ \\

\textbf{C$^2$RL} &
\hlg{世界上没有} \hlr{不} \hlg{后悔} \hlr{的作弊} \newline
\textit{\hlg{There is no}\hlr{such thing as cheating without}\hlg{regret.}} &
B-4: 40.35 \newline
R-L: 70.27 \\

\textbf{Ours} &
\hlg{世界上没有后悔药} \newline
\textit{\hlg{There is no medicine for regret in the world.}} &
B-4: 100.00 \newline
R-L: 100.00 \\
\midrule

\textbf{Reference} &
不要在家里看电视，出去踢足球 \newline
\textit{(Don't stay home watching TV, go out and play football.)} &
~ \\

\textbf{C$^2$RL} &
\hlp{不要}\hlg{在家看电视}\hly{不去了}\hlp{出去}\hlg{踢足球}\newline
\textit{\hlp{Don't}\hly{staying}\hlg{home and watching TV}\hlr{instead of}  \hly{playing}\hlg{football.}} &
B-4: 40.39 \newline
R-L: 76.03 \\

\textbf{Ours} &
\hlg{不要在家里看电视，}\hly{去}\hlg{踢足球}\newline
\textit{\hlg{Don't stay home watching TV,}\hly{go}\hlg{play football.}} &
B-4: 78.82 \newline
R-L: 96.03 \\
\midrule

\textbf{Reference} &
售货员在热情地向顾客介绍商品 \newline
\textit{(The salesperson is enthusiastically introducing the products to the customer.)} &
~ \\

\textbf{C$^2$RL} &
\hlr{服务员不以冷淡的态度对待向}\hlg{顾客}\hlr{提供}\hlg{商品}\newline
\textit{\hlr{Waiters should not treat}\hlg{customers}\hlr{with indifference}
\hlr{when serving}\hlg{goods.}} &
B-4: 0.00 \newline
R-L: 30.47 \\

\textbf{Ours} &
\hly{商品员}\hlg{在热情地向顾客介绍商品} \newline
\textit{\hly{The sales clerk}\hlg{enthusiastically}\hly{introduced}\hlg{the}
\hlg{products to the customer.}
} &
B-4: 69.81 \newline
R-L: 81.26 \\
\bottomrule
\end{tabular}

\vspace{-2mm}
\caption{\small
Qualitative and quantitative comparison between C$^2$RL and our method on the CSL-Daily dataset.
English translations are obtained via \textit{Google Translate}. Text segments highlighted in \hlg{green} denote exact matches with the reference, \hly{yellow} indicates semantically similar expressions, \hlp{purple} marks omissions, and \hlr{red} denotes incorrect meanings.
B-4: BLEU-4, R-L: ROUGE-L
}
\vspace{-2mm}
\label{tab:qualitative-results-csl-daily}
\end{table}
\end{CJK*}

\paragraph{Reward Learning Curve.}
We evaluate the checkpoints of the RFT stage every 100 steps to plot the learning curve, as shown in Fig.~\ref{fig:reward}. The SLT-GRPO model consistently outperforms SLT-SFT across all checkpoints, showing continuous performance growth from 0 up to approximately 4500 training steps.

\vspace{-4mm}
\paragraph{Case Study.}
We visualize the translation outputs of our method and C$^2$RL on three randomly selected CSL-Daily samples. As shown in Table~\ref{tab:qualitative-results-csl-daily}, both models generate reasonable translations, yet C$^2$RL shows clear errors and omissions. Our method achieves higher lexical overlap and closer semantic alignment with the references, resulting in more accurate sentence-level correspondence. Even when wording differs, the intended meanings remain consistent. These findings indicate that richer sign representations allow the LLM to capture finer semantic cues, while GRPO-based optimization further improves translation fidelity and completeness. Additional examples are included in the supplementary material.

\section{Conclusion}
This paper presents RVLF, a unified three-stage framework for sign language translation that integrates cross-lingual visual-language pre-training, instruction-tuned fine-tuning, and GRPO-based reinforcement optimization. The proposed approach enhances sign representations and aligns sentence-level semantics, resulting in significant improvements in translation fidelity and completeness. Experiments on multiple benchmark datasets show that RVLF achieves state-of-the-art performance, surpassing even models pre-trained on large-scale external sign language corpora.

\vspace{-4mm}
\paragraph{Limitations and Future Work.}
RVLF performs well without external pre-training, but its results are limited by the size and diversity of current benchmarks. Due to computational constraints, LoRA was used during LLM fine-tuning, and the model's full potential remains unexplored. Future work will focus on scaling the framework with larger, real-world sign language datasets.
{
    \small
    \bibliographystyle{ieeenat_fullname}
    \bibliography{main}
}
\clearpage
\setcounter{page}{1}
\maketitlesupplementary

To enable a thorough understanding of RVLF, this supplementary material presents more implementation details, further analyses, and more case studies. The core implementation code will be provided in the Supplementary Materials submitted alongside this file.

\section{More Implementation Details}\label{sec:More Details}

\subsection{Keypoints extraction.}
We follow the keypoint extraction strategy in Cosign \cite{cosign} and use the off-the-shelf estimator MMPose to obtain full-body keypoints from each frame of the sign-language videos. We select keypoints related to sign representation, including 9 body points, 21 points for each hand, 8 mouth points, and 18 facial points, giving a total of 77 keypoints. More details can be found in the Cosign \cite{cosign}.

\subsection{DINOv2 Feature extraction.}
We first use full-body keypoints to detect the bounding boxes of the hands and face, then crop these regions and resize them to 224×224. If the detector fails to locate the hands or face in a given frame, we replace that frame with the most recent frame where detection succeeded. The cropped regions are then fed into a pre-trained DINOv2 model (facebook/dinov2-base) to extract semantically rich feature sequences.

\subsection{Training Setting.}
In this section, we provide as many training details of RVLF as possible, including the computing infrastructure and the training configurations for each stage.
\paragraph{Computing Infrastructure}
The entire experimental process was conducted on a computational infrastructure comprising an Intel(R) Xeon(R) Gold 6226R CPU @ 2.90GHz, 4 Tesla V100S GPU with 32GB of VRAM, running the Ubuntu 20.04 operating system, with Python 3.10, CUDA 11.8, transformers 4.51.3 and PyTorch 2.1.2.

\begin{table}[t]
    \centering
    \scriptsize
    \begin{tabular}{l c | c c}
        \toprule
        \multicolumn{2}{c|}{Training settings} & \multicolumn{2}{c}{Model settings} \\
        \midrule
        Config          & Value         & Config    & Value \\
        \midrule
        epoch           & 200           & MHA Number        & 8           \\
        Optimizer       & SGD           & Layer Number      & 3         \\
        Learning Rate   & 0.01          & Hidden Size       & 512        \\
        Weight Decay    & 0.001         & FF Dimension      & 2048       \\
        Scheduler       & Cosine        & Beam Search Width & 5       \\
        Cycle Decay     & 1.0           & Vocab Size        & 6201        \\
        Batch Size      & 8             & Max Length        & 300        \\
        Criterion      & CrossEntropy  & Con Dimension     & 1024         \\
        \bottomrule
    \end{tabular}
    \vspace{-2mm}
    \caption{\small Settings for pre-training stage. FF Dimension : feed-forward dimension; Con Dimension : Contrastive Dimension.}
    \label{tab:stage1}
    \vspace{-2mm}
\end{table}%

\begin{table}[t]
    \centering
    \scriptsize
    \begin{tabular}{l c | c c}
        \toprule
        \multicolumn{2}{c|}{Training settings} & \multicolumn{2}{c}{Model settings} \\
        \midrule
        Config          & Value         & Config    & Value \\
        \midrule
        Epoch           & 40            & Projector Modules & mlp + gelu     \\
        Batch Size      & 4             & MLP Dim           & 512,4096,4096 \\
        Optimizer       & AdamW         & LLM               & Qwen3-8B   \\
        LR              & $2 \times 10 ^{-4}$          & Target Modules    & q\_proj, v\_proj  \\
        Projector LR    & $2 \times 10 ^{-5}$          & Rank              & 16        \\
        Weight Decay    & 0             & Scaling Factor    & 32       \\
        Scheduler       & Cosine        & LoRA Dropout      & 0.3       \\
        Cycle Decay     & 1.0           & Max Length        & 300        \\
        Criterion      & CrossEntropy  & Beam Search Width & 5         \\
        \bottomrule
    \end{tabular}
    \vspace{-2mm}
    \caption{\small Settings for supervised fine-tuning stage. LR : Learning Rate; q\_proj : query projection matrices; v\_proj : value projection matrices.}
    \label{tab:vitung}
    \vspace{-2mm}
\end{table}%

\paragraph{Settings on Pre-training Stage}
The settings for the pre-training stage are summarized in Table \ref{tab:stage1}. This stage involves training for 200 epochs using the SGD optimizer with a learning rate of 0.01 and a weight decay of 0.001. A cosine learning rate scheduler is applied, with a warm-up period of 20 epochs, which corresponds to 10\% of the total training epochs. The batch size is set to 8, and the cross-entropy loss is used as the training objective. In the pre-training stage, both the text encoder and decoder are built by reinitializing the first three layers of the mBART model. In contrast, the skeleton encoder is implemented as a mBART encoder with three layers. The text encoder, decoder, and skeleton encoder all follow the same configuration, where each Transformer block consists of an 8-head multi-head self-attention mechanism, a hidden size of 512, and a feed-forward network with a dimensionality of 2048. Following \cite{GFSLT-VLP}, we reduce the original vocabulary to a task-specific subset, resulting in a final vocabulary size of 6,201 (for CSL-Daily Dataset). The maximum input sequence length is limited to 300, as explained in Section \ref{sec:why_300}. For contrastive learning, we fix the feature dimension of both vision and text inputs to 1024. During text decoding, we use a beam width of 5.

\begin{figure*}[!t]
\centering
\includegraphics[width=\linewidth]{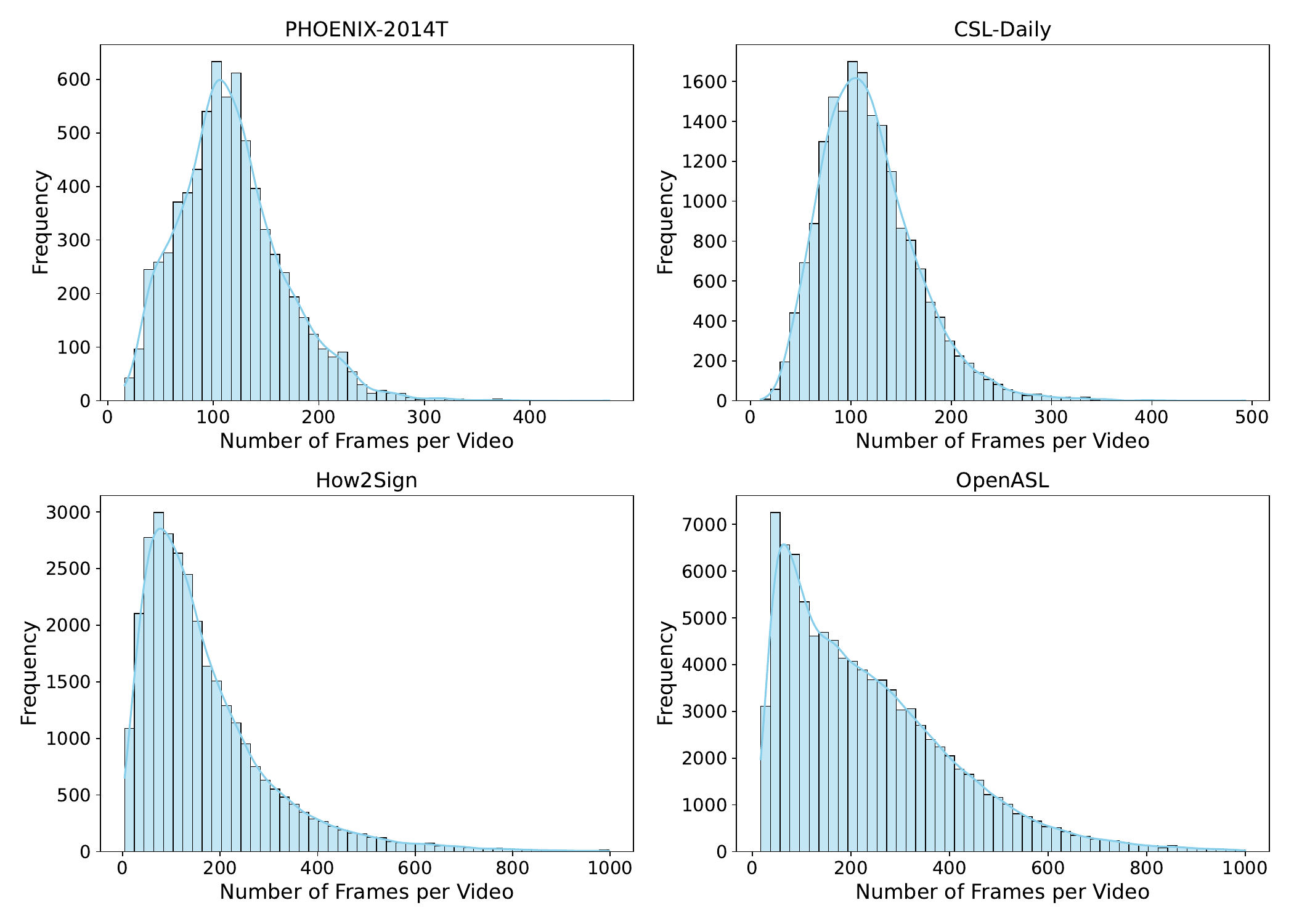}
\vspace{-2mm}
\caption{\small
Visualization of the input length distribution of sign language videos in the training sets of four datasets: PHOENIX-2014T, CSL-Daily, PHOENIX-2014T How2Sign, and OpenASL.
Due to the significant long-tail phenomenon in the frame count distributions of How2Sign and OpenASL, only samples with a frame count of 1000 or fewer are included in the visualization to improve readability, while still covering the vast majority of samples in both datasets.
}
\label{fig:data_ana}
\vspace{-2mm}
\end{figure*}
\begin{table}[t]
    \centering
    \scriptsize
    \begin{tabular}{l c | c c}
        \toprule
        \multicolumn{2}{c|}{Training settings} & \multicolumn{2}{c}{Model settings} \\
        \midrule
        Config          & Value         & Config    & Value \\
        \midrule
        Epoch           & 2             & LLM               & Qwen3-8B \\
        Batch Size      & 8             & Target Modules    & q\_proj, v\_proj   \\
        Optimizer       & AdamW         & Rank              & 64  \\
        LR              & $2 \times 10 ^{-5}$          & Scaling Factor    & 128        \\
        Responses       & 8             & LoRA Dropout      & 0.05       \\
        Weight Decay    & 0             & Max Length        & 300       \\
        Criterion       & BELU \& ROUGE & Beam Search Width & 5        \\
        \bottomrule
    \end{tabular}
    \vspace{-2mm}
    \caption{\small Settings for reinforcement fine-tuning stage.}
    \label{tab:RFT}
    \vspace{-2mm}
\end{table}%

\paragraph{Settings on Supervised Fine-Tuning Stage}
The training settings for the supervised fine-tuning stage are shown in Table \ref{tab:vitung}. The model is trained for 40 epochs using the AdamW optimizer. The LLM parameters use a learning rate of $2 \times 10 ^{-4}$, while the multimodal projector uses a learning rate of $2 \times 10 ^{-5}$. Weight decay is set to 0. A cosine learning rate schedule with 8 warmup epochs is applied. The batch size is 4, and cross-entropy loss is used as the training objective. Following LLaVA \cite{llava}, we adopt a two-layer MLP projector with GELU activation, which expands the 512-dimensional visual feature to 4096 dimensions and applies another 4096-dimensional linear layer to produce visual embeddings compatible with the language model. Qwen3-8B serves as the base language model. To enable parameter-efficient training, we apply LoRA to the query and value projection matrices in the self-attention layers, using rank 16, a scaling factor of 32, and a dropout rate of 0.3. The instruction template is set as: \textit{“A conversation between a deaf person and an AI assistant. The assistant can generate accurate and fluent translated text based on provided sign language visual information.”} The maximum input sequence length is 300, and the motivation is discussed in Sec. \ref{sec:why_300}. During inference, we use deterministic beam search with a beam width of 5 and disable sampling.

\paragraph{Settings on Reinforcement Fine-Tuning Stage}
After merging the LoRA parameters from the SFT stage into the base model, we obtain the SLT-SFT model. The training settings for the reinforcement fine-tuning stage are summarized in Table \ref{tab:RFT}. In this stage, we add new LoRA modules to the query and value projection matrices in the self-attention layers, using rank 64, a scaling factor of 128, and a dropout rate of 0.05. We use the AdamW optimizer with a learning rate of $2 \times 10 ^{-5}$, train for two epochs, and use a batch size of 8. Following the \cite{grpo} setting and using the same prompts as in the SFT stage, we sample 8 candidate responses for each input. These responses are grouped to compute relative rewards based on BLEU-4 and ROUGE-L and to estimate advantages for policy updates. As in the SFT stage, the maximum input sequence length is set to 300. During inference, we use beam search with a beam width of 5. A practical detail is that some sign language datasets contain very short sentences. To reduce unstable BLEU-4 scores that may interrupt training, we append additional punctuation to the end of these sentences.

\begin{table}[!t]\scriptsize
\centering
\begin{tabular}{lcccc}
\toprule
\textbf{Frame Length} & 
\textbf{PHOENIX14T} & 
\textbf{CSL-Daily} & 
\textbf{How2Sign} & 
\textbf{OpenASL} \\
\midrule

${>}300$   & 23    & 90     & 4338   & 29172 \\
${>}500$   & 0     & 0      & 1180   & 8226  \\
${>}800$   & 0     & 0      & 258    & 1138  \\
${>}1000$  & 0     & 0      & 140    & 362   \\
\midrule

Total Samples &
7096 &
18401 &
31064 &
95888 \\

\bottomrule
\end{tabular}

\vspace{-2mm}
\caption{\small 
Comparison of training samples exceeding different frame-length thresholds across four large-scale sign language datasets: PHOENIX14T, CSL-Daily, How2Sign, and OpenASL.
}
\vspace{-2mm}
\label{tab:max_len}
\end{table}

\begin{figure}[!t]
\centering
\includegraphics[width=\linewidth]{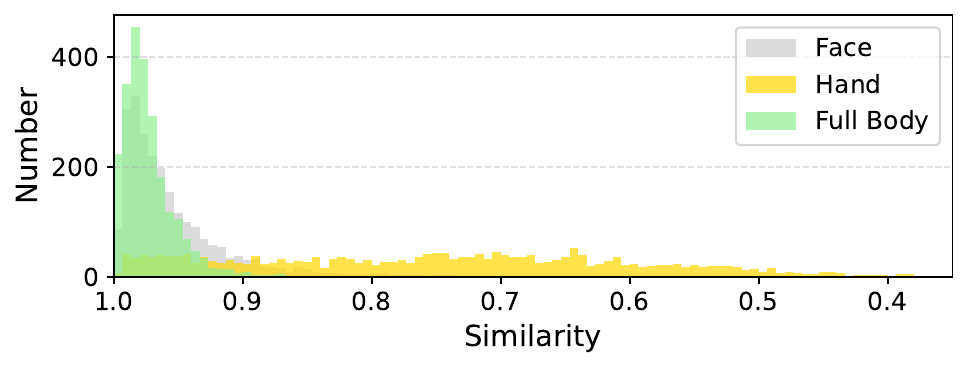}
\vspace{-7mm}
\caption{\small Stacked histogram of frame-to-frame similarity for full body, face, and hand regions.}
\label{fig:vision_diff}
\vspace{-3mm}
\end{figure}

\section{Further Analyses} \label{sec:Further Analyses}

\subsection{Dataset Input Sequence Length Analysis} \label{sec:why_300}
The sign language datasets exhibit substantial variation in sequence length. As shown in Figures~\ref{fig:data_ana} and Table~\ref{tab:max_len}, How2Sign and OpenASL contain a markedly higher proportion of long sequences than PHOENIX14T and CSL-Daily, reflecting greater linguistic complexity. Due to GPU memory limits and heterogeneous length distributions, all sequences are truncated to 300 frames during training. 
In contrast, the VAP paper employs NVIDIA RTX A800 GPUs, which likely support longer sequences and require less truncation. This difference may partially explain the larger performance gains reported on their benchmarks relative to the more modest improvements observed on How2Sign and OpenASL. The limited improvement on PHOENIX14T is more plausibly due to the already strong performance of existing methods on this dataset and the comparatively smaller amount of German-language training data available in Qwen3.

\subsection{Effect of Visual Representations.}

\paragraph{Visual Region}
The analysis of frame-to-frame cosine similarity for full-body, face, and hand features reveals clear temporal differences. Cosine similarity is computed between adjacent frames, and the distribution is shown in Figure~\ref{fig:vision_diff}. Hand and face features show greater variation compared to full-body features, indicating that they contain rich temporal cues that are crucial for modeling sign transitions. Full-body features, on the other hand, remain relatively stable because static regions and background elements dominate the representation. These findings suggest that the hand and face regions provide more informative and discriminative visual signals, which are essential for temporal modeling in sign language understanding.

\definecolor{mygreen}{RGB}{198,239,206}
\definecolor{myyellow}{RGB}{255,242,204}
\definecolor{mypurple}{RGB}{218,210,233}
\definecolor{myred}{RGB}{255,204,204}
\newcommand{\hlg}[1]{\colorbox{mygreen}{#1}}
\newcommand{\hly}[1]{\colorbox{myyellow}{#1}}
\newcommand{\hlp}[1]{\colorbox{mypurple}{#1}}
\newcommand{\hlr}[1]{\colorbox{myred}{#1}}

\begin{table*}[!t]
\centering
\scriptsize
\setlength{\tabcolsep}{6pt}
\resizebox{\linewidth}{!}{
\begin{tabular}{@{}p{1.2cm}|p{11.2cm}|p{3cm}@{}}
\toprule
\textbf{Source} & \textbf{Sentence} & \textbf{Objective Metrics} \\
\midrule

\textbf{Reference} &
liebe zuschauer guten abend \newline
\textit{(Dear viewers, good evening)} &
~ \\

\textbf{C$^2$RL} &
\hlg{liebe zuschauer}\hlp{guten abend} \newline
\textit{(\hlg{Dear viewers}\hlp{,good evening})} &
BLEU-4: 0.00 \newline
ROUGE\_L: 67.11 \\

\textbf{Ours} &
\hlg{liebe zuschauer guten abend} \newline
\textit{(\hlg{Dear viewers, good evening})} &
BLEU-4: 100.00 \newline
ROUGE\_L: 100.00 \\
\midrule

\textbf{Reference} &
am sonntag ziehen von nordwesten wieder schauer und gewitter heran \newline
\textit{(On Sunday, showers and thunderstorms will approach again from the northwest)} &
~ \\

\textbf{C$^2$RL} &
\hlg{am sonntag}\hlr{breiten sich von westen}\hlg{wieder schauer und gewitter}
\hlr{aus}\hlg{heran} \newline
\textit{(\hlg{On Sunday, showers and thunderstorms will}\hlr{spread from the west})} &
BLEU-4: 0.00 \newline
ROUGE\_L: 63.64 \\

\textbf{Ours} &
\hlg{am sonntag ziehen}\hly{dann}\hlg{von nordwesten schauer und gewittern heran} \newline
\textit{(\hlg{On Sunday, showers and thunderstorms will}\hly{then}\hlg{approach from the northwest})} &
BLEU-4: 51.57 \newline
ROUGE\_L: 90.91 \\
\midrule

\textbf{Reference} &
in der nacht muss vor allem in der nordwesthälfte mit schauern und gewittern gerechnet werden die heftig ausfallen können \newline
\textit{(During the night, showers and thunderstorms are to be expected, especially in the northwestern half, which can be severe)} &
~ \\

\textbf{C$^2$RL} &
\hlg{in der nacht muss vor allem}\hly{im nordwesten}\hlg{mit}\hlr{einzelnen}\hlg{schauern und gewittern gerechnet werden die}
\hly{mitunter kräftig sein}\hlg{können}\newline
\textit{(\hlg{During the night, isolated showers and thunderstorms are to be expected, especially in the}\hly{northwest}\hlp{, which}
\hlp{can be severe}) } &
BLEU-4: 48.46 \newline
ROUGE\_L: 73.08 \\

\textbf{Ours} &
\hlg{in der nacht muss vor allem in der nordwesthälfte mit schauern und gewittern gerechnet werden die}
\hly{mitunter kräftig}\hlg{ausfallen können}\newline
\textit{(\hlg{During the night, showers and thunderstorms are to be expected, especially in the northwestern half, which}
\hly{can sometimes be heavy})} &
BLEU-4: 81.37 \newline
ROUGE\_L: 92.57 \\
\bottomrule

\end{tabular}
}
\caption{\small
Qualitative and quantitative comparison between C$^2$RL and our method on the \textbf{PHOENIX14T} dataset. English translations are obtained via \textit{Google Translate}. Text segments highlighted in \hlg{green} denote exact matches with the reference, \hly{yellow} indicates semantically similar expressions, \hlp{purple} marks omissions, and \hlr{red} denotes incorrect meanings.
}
\label{tab:qualitative-results-p14t}
\end{table*}

\begin{figure}[!t]
    \centering
    \includegraphics[width=0.5\linewidth]{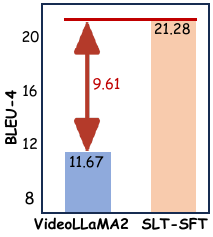}
    \vspace{-2mm}
    \caption{Performance comparison between directly fine-tuning VideoLLaMA2 and the SLT-SFT model.}
    \label{fig:VideoLLaMA2}
    \vspace{-3mm}
\end{figure}

\paragraph{Visual Encoder}

\begin{figure}[t]
    \centering
    \includegraphics[width=0.9\linewidth]{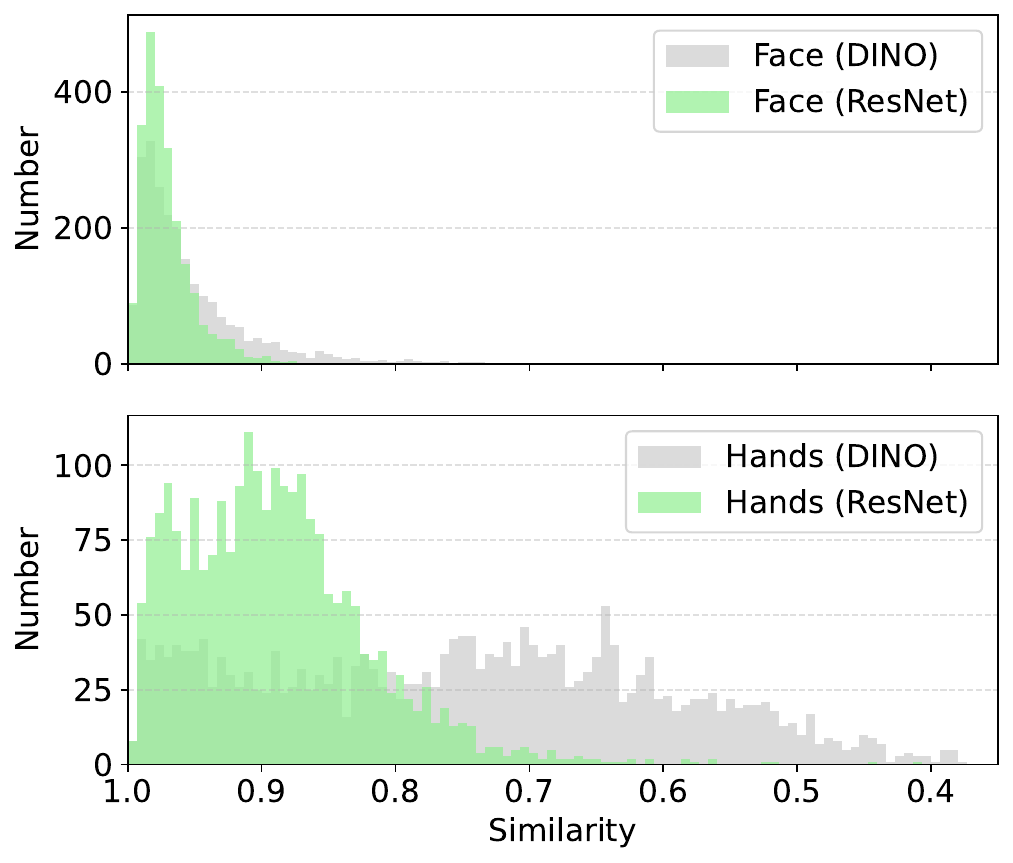}
    \vspace{-2mm}
    \caption{Analyzing the impact of vision distribution. Cosine similarity distribution map between adjacent frames of ResNet-18 and DINOv2. Twenty different samples were randomly selected from the CSL-Daily dataset.}
    \label{fig:vis_dist}
\end{figure}

To investigate the capabilities of different pretrained visual encoders on the SLT task, we examine the feature distributions output by various visual encoder backbones. Specifically, we compute the inter-frame cosine similarity of face and hand region features output by different encoders (i.e., DINOv2 and ResNet-18). For the ResNet-18 baseline, we follow \cite{GFSLT-VLP} and employ a ResNet model pretrained on ImageNet. As shown in Fig.~\ref{fig:vis_dist}, features extracted by ResNet-18 exhibit high redundancy, with similarity scores densely concentrated around 1.0, indicating its inability to capture subtle yet crucial temporal variations in sign language. By contrast, DINOv2 yields a broader and more diverse similarity distribution, particularly for hand features, indicating that DINOv2 can provide informative sign representations thanks to its strong vision perception.

\subsection{Directly Fine-Tuning a Video LLM}

\begin{table*}[!t]
\centering
\scriptsize
\setlength{\tabcolsep}{6pt}
\resizebox{\linewidth}{!}{
\begin{tabular}{@{}p{1.2cm}|p{11.2cm}|p{3cm}@{}}
\toprule
\textbf{Source} & \textbf{Sentence} & \textbf{Objective Metrics} \\
\midrule

\textbf{Reference} &
i'm here telling you how to wash your jeans &
~ \\

\textbf{C$^2$RL} &
\hlg{i'm here}\hly{to talk to you about}\hlg{how to wash your}\hlr{clothes} &
BLEU-4: 26.20 \newline
ROUGE\_L: 64.10 \\

\textbf{Ours} &
\hlg{i'm here}\hly{to tell you}\hlg{how to wash your jeans} &
BLEU-4: 58.14 \newline
ROUGE\_L: 83.75 \\
\midrule

\textbf{Reference} &
you can take this forward and back, you can take it in a circle, you can take it in a lot of different directions. &
~ \\

\textbf{C$^2$RL} &
\hlg{you can}\hlr{put it on the floor},\hlr{you can put it around the house},\hlg{you can}\hlr{do} \hlg{different} \hlr{things} &
BLEU-4: 0.00 \newline
ROUGE\_L: 36.24 \\

\textbf{Ours} &
\hlg{you can take this forward and back,}\hlg{you can take it} \hlr{sideways}
\hlg{,you can take it in a lot of different directions}. &
BLEU-4: 79.67 \newline
ROUGE\_L: 90.96 \\
\midrule

\textbf{Reference} &
so, i'm going to turn on my sequencer and i'm just going to press play and you can just go to each one and just hear a different presets &
~ \\

\textbf{C$^2$RL} &
\hlr{i'm hooping, my note cards,}\hly{i'm going to}\hlr{these players as fast as you can, and you're just going to follow one }
\hlr{and}\hly{hear the difference}\hlr{between the note cards} &
BLEU-4: 0.00 \newline
ROUGE\_L: 36.24 \\

\textbf{Ours} &
\hlg{so}\hlr{i changed my settings and}\hlg{i'm just going to press play and you can just go to each one and hear the different}
\hlg{presets} &
BLEU-4: 79.67 \newline
ROUGE\_L: 90.96 \\
\bottomrule

\end{tabular}
}
\caption{\small
Qualitative and quantitative comparison between C$^2$RL and our method on the \textbf{How2Sign} dataset. Text segments highlighted in \hlg{green} denote exact matches with the reference, \hly{yellow} indicates semantically similar expressions, \hlp{purple} marks omissions, and \hlr{red} denotes incorrect meanings.
}
\label{tab:qualitative-results-how2sign}
\end{table*}

\begin{table*}[!t]
\centering
\scriptsize
\setlength{\tabcolsep}{6pt}
\resizebox{\linewidth}{!}{
\begin{tabular}{@{}p{1.2cm}|p{11.2cm}|p{3cm}@{}}
\toprule
\textbf{Source} & \textbf{Sentence} & \textbf{Objective Metrics} \\
\midrule

\textbf{Reference} &
Port Authority said the girl was playing with her grandfather in a dining hall &
~ \\

\textbf{C$^2$RL} &
\hlr{Grant's}\hlg{authorities said}\hlr{he is a} \hlg{girl}\hlr{and that he}\hly{plays with his} \hlg{grandfather}\hlr{at a nursing home} \newline
&
BLEU-4: 0.00 \newline
ROUGE\_L: 30.32 \\

\textbf{Ours} &
\hly{The} \hlg{Port Authority said the girl was playing with her grandfather in the dining hall} &
BLEU-4: 74.87 \newline
ROUGE\_L: 89.44 \\
\midrule

\textbf{Reference} &
Bernie Sanders won big in Nevada's  on Saturday with 47\% of the popular vote &
~ \\

\textbf{C$^2$RL} &
\hlg{Bernie Sanders won}\hlr{the primaries in}\hlg{Nevada on Saturday}\hlr{and his popular caucuses on Saturday} &
BLEU-4: 0.00 \newline
ROUGE\_L: 46.67 \\

\textbf{Ours} &
\hlg{Bernie Sanders won}\hlp{big in}\hlg{Nevada's caucuses on Saturday with 47\% of the popular vote} &
BLEU-4: 72.98 \newline
ROUGE\_L: 91.92 \\
\midrule

\textbf{Reference} &
Thursday marks exactly six months since the World Health Organization declared Covid-19 a public health emergency &
~ \\

\textbf{C$^2$RL} &
\hlr{As of Thursday morning}\hlg{the World Health Organization}\hlr{(WHO) has been}\hlg{declared}\hlr{a "coronavirus crisis."} &
BLEU-4: 17.70 \newline
ROUGE\_L: 45.07 \\

\textbf{Ours} &
\hlg{Thursday marks}\hlp{exactly}\hlg{six months since the World Health Organization declared Covid-19 a public health}
\hlg{emergency} &
BLEU-4: 84.15 \newline
ROUGE\_L: 96.57 \\
\bottomrule

\end{tabular}
}
\caption{\small
Qualitative and quantitative comparison between C$^2$RL and our method on the \textbf{OpenASL} dataset. Text segments highlighted in \hlg{green} denote exact matches with the reference, \hly{yellow} indicates semantically similar expressions, \hlp{purple} marks omissions, and \hlr{red} denotes incorrect meanings.
}
\label{tab:qualitative-results-openasl}
\end{table*}


Collecting large-scale sign language datasets is difficult because annotation is expensive and skilled signers are limited. This leads to a clear gap between visual and textual modalities. Video LLMs provide strong temporal modeling and natural language generation abilities, which suggests that fine-tuning them with only a small number of sign--text pairs could support effective cross-modal alignment while reducing both data requirements and training cost. As a result, adapting powerful Video LLMs for sign language translation is a promising direction. To ensure a fair comparison and isolate the contribution of the model architecture, we strictly follow the official training configuration of VideoLLaMA2: all training settings and hyperparameters remain unchanged, and only the video data is replaced with sign-language datasets. However, as shown in Fig.~\ref{fig:VideoLLaMA2}, our earlier experiments reveal a large performance gap between directly fine-tuning VideoLLaMA2 (7B)~\cite{videollama2} on sign--text pairs and the SLT-SFT model. A main reason is that sign language semantics are more structured and abstract than general video content, and the visual encoder of VideoLLaMA2 fails to learn discriminative sign representations. In addition, the linguistic granularity required for SLT is substantially higher than that for generic video understanding, making it difficult for a general-purpose video encoder to capture subtle phonological and morphological cues in sign videos. Although our analysis is based on a relatively simple experimental setting, exploring Video LLMs with stronger visual backbones and larger parameter scales remains a promising direction for future work.

\section{More Case Studys}\label{sec:case_study}
We conducted qualitative and quantitative analyses of the texts generated by RVLF and C$^2$RL using translation samples from the PHOENIX14T, How2Sign, and OpenASL test sets, as shown in Tables~\ref{tab:qualitative-results-p14t}, \ref{tab:qualitative-results-how2sign}, and \ref{tab:qualitative-results-openasl}. The outputs from RVLF contain a larger number of words that match the ground-truth references, and their overall semantics are more consistent with the reference texts. This improvement is mainly due to the richer sign language representations produced by RVLF, which allow the model to capture semantic information more effectively. Additionally, RVLF benefits from a more robust feature extraction process, enabling it to better handle variations in sign language data. In addition, RVLF introduces a sentence-level reinforcement optimization strategy that helps maintain stronger semantic consistency and improves translation accuracy.
\end{document}